\documentclass[11pt]{article}

\usepackage[margin=1in]{geometry}
\usepackage[T1]{fontenc}
\usepackage[utf8]{inputenc}
\usepackage{lmodern}
\usepackage{setspace}
\usepackage{hyperref}
\usepackage{csquotes}
\usepackage[numbers,sort&compress]{natbib}
\usepackage{tabularx}
\hypersetup{
  pdftitle={Berkeley and Heiserman as an Unexhausted Architecture for Embodied Machine Intelligence},
  pdfauthor={Christopher A. Tucker},
  pdfsubject={Embodied machine intelligence, symbolic logic, and robotic control architecture},
  pdfkeywords={embodied AI, machine intelligence, symbolic logic, robotics, cybernetics, Berkeley, Heiserman}
}

\title{Berkeley and Heiserman as an Unexhausted Architecture for\\
Embodied Machine Intelligence}
\author{Christopher A. Tucker\\
\small \texttt{cartheur@pm.me}\\
\small \href{https://orcid.org/0000-0002-0172-7258}{ORCID: 0000-0002-0172-7258}}
\date{}

\begin{document}

\maketitle

\begin{abstract}
Edmund C. Berkeley is usually remembered as a mediator between symbolic logic and early computing, yet that standard description understates the scope of his work. This paper argues for a stronger reading: Berkeley should also be understood as an early theorist of embodied machine intelligence. Across Berkeley's major writings on symbolic logic, machine intelligence, living robots, and Squee, intelligence appears not as disembodied symbol manipulation alone but as the organized coordination of sensing, storage, calculation, control, state, and action in physically realized machines. The paper's first contribution is interpretive: it reconstructs Berkeley as a thinker of machine architecture, temporally extended behavior, and environment-coupled control. Its second contribution is comparative: it reads Berkeley alongside David L. Heiserman to recover a shared descriptive scheme centered on sensing, state or memory, control, action, and adaptation. Its third contribution is critical: it uses that scheme to assess current embodied-AI discourse. The broader claim is that contemporary LLM-centered robotics often demonstrates impressive capability without an equally explicit account of persistence, recoverability, maintenance, and structured modification of conduct through experience.
\end{abstract}

\onehalfspacing

\section{Introduction}

Contemporary AI and robotics research often oscillates between large-scale statistical learning, abstract planning systems, and highly specialized control solutions. In that environment, behavior-organizing approaches rooted in explicit states, logical conditions, hardware-near control, and modular sensorimotor structure can be dismissed too quickly as dated. This paper begins from the opposite suspicion: that these design commitments remain technically coherent and still experimentally available for robotic cognition. The question is not whether twentieth-century machine theories should be repeated unchanged, but whether certain older design commitments still describe a usable architecture for embodied intelligence. Current discussion often overlooks that architecture because it is distributed across logic, control, robotics, and early machine-intelligence writing rather than consolidated under a single contemporary label.

That neglect shapes how contemporary progress is narrated. Recent robotics work built around large multimodal models is often presented as if the serious problem of embodied intelligence had only now become visible: the machine must perceive, reason, plan, adapt, and control a whole body in a changing world \citep{googledeepmind2025geminirobotics,googledeepmind2026geminier}. Those achievements are real and should not be minimized. But the public framing around them can encourage historical amnesia, as though coupling perception, state, action, world-modeling, and adaptive correction were concepts born with vision-language-action systems or contemporary embodied reasoning models. One purpose of the present paper is to resist that flattening. Berkeley, Ashby, Walter, and later Heiserman were already working on closely related questions, often in cruder hardware and with a smaller mathematical toolkit, but with a level of architectural explicitness that remains instructive. If current systems sometimes appear to succeed by scale, data, and iterative trial instead of by an articulated behavior vocabulary, that is not a reason to dismiss them. It is a reason to recover older design languages that can help us describe what such systems are actually doing.

Edmund C. Berkeley is commonly placed in the history of symbolic logic, early computing, and machine thought. That placement is justified. \emph{Symbolic Logic and Intelligent Machines} is overtly concerned with Boolean algebra, logical relations, class reasoning, and the design of information-handling machinery \citep{berkeley1959symbolic}. Yet this standard placement narrows Berkeley's significance. He was interested in the formal calculability of statements and classes, but he was also interested in how reason-like competence could be built into machinery that senses, stores, decides, and acts. That broader concern is easy to miss if Berkeley is read only through the lens of formal logic. It becomes clearer when the same book is read as a document about machine organization: what kinds of parts an intelligent machine has, how those parts relate, how control is represented, and how behavior unfolds through states and events.

The present argument is therefore developed across Berkeley's major writings on symbolic logic, machine intelligence, living robots, and Squee, treated here as a dispersed but coherent corpus of machine-intelligence thought rather than as isolated curiosities.

The aim of this paper is to argue that Berkeley should be read as a theorist of embodied machine intelligence, not just as a symbolic logician of machines. He is not a full theorist of adaptive AI in the Heiserman sense, but neither is he adequately understood as a purely formal or disembodied thinker. His work gives a logically rigorous account of how physically organized machines can take in information, retain it, operate on it, and guide their own activity over time \citep[pp.~66--70]{berkeley1959symbolic}. When read alongside Heiserman's creature-based machine intelligence, Berkeley's work appears as an early account of problems that remain active in robotics. The paper therefore asks the reader to shift levels of attention: away from Berkeley as a contributor to symbolic notation and toward Berkeley as a thinker of machine architecture, behavior, and temporally extended control. It also asks the reader to understand the paper's later robotic material in the right order of importance. The AIBO section is not the paper's center of gravity and not its chief novelty claim. Its role is subordinate and demonstrative: it shows that the reconstructed Berkeley--Heiserman vocabulary remains useful for describing preserved embodied behavior.

The deeper tension of the paper follows from that shift. The present argument is historical and taxonomic, but it also bears on contemporary embodied robotics, especially on research cultures that increasingly treat large language models and multimodal foundation models as if they were approaching a sufficient account of robot intelligence once they can interpret instructions, integrate perception, and drive end-to-end behavior. The paper does not deny the importance of those developments. It treats them as evidence that machine intelligence has once again become an engineering problem at the level of whole embodied systems. But it argues that such systems are often described in a way that is thinner than their achievement would seem to require. They may display impressive competence without yet providing an equally explicit account of persistence, recoverability, world-coupling, internally organized state, maintenance, and the structured modification of conduct through experience. The Berkeley--Heiserman line matters here because it supplies terms for asking whether a robotic system is effective and explicitly organized as a controlled embodied machine that persists, recovers, and changes through time. The paper's claim is therefore stronger than historical recovery and narrower than a blanket rejection of present methods: current LLM-centered robotics may be powerful and important without being mistaken in principle, yet still remain undertheorized as accounts of living, controlled embodied intelligence.

Put in the most compact form, this paper makes three contributions. First, it offers a reinterpretation of Berkeley as a serious thinker of embodied control architecture, not just as a symbolic logician of machines. Second, it reconstructs a Berkeley--Heiserman descriptive vocabulary in which intelligence is organized through explicit relations among sensing, state or memory, control, action, and adaptation. Third, it uses that vocabulary to examine present embodied-AI discourse, where capability is often more visible than the underlying organization of machine conduct.

\begin{center}
\small
\setlength{\tabcolsep}{5pt}
\begin{tabularx}{\textwidth}{@{}>{\raggedright\arraybackslash}p{0.18\textwidth}>{\raggedright\arraybackslash}X>{\raggedright\arraybackslash}X@{}}
\hline
Focus & Main claim & Function in the paper \\
\hline
Berkeley & Berkeley describes intelligent machinery as an organized relation among sensing, storage, calculation, control, states, events, and action. & Reinterprets Berkeley as an architect of embodied machine intelligence, not just a symbolic logician. \\
Heiserman & Heiserman extends this design space toward memory-guided response, generalization, and adaptive creature intelligence. & Supplies the comparative developmental layer that clarifies Berkeley's reach and limits. \\
Present robotics & Contemporary embodied AI often shows strong capability without an equally explicit vocabulary of persistence, recoverability, maintenance, and organized revision. & Uses the reconstructed Berkeley--Heiserman vocabulary as a present-day comparison, not just a historical association. \\
\hline
\end{tabularx}
\end{center}

\section{Question and Working Thesis}

The guiding question is how Berkeley's \emph{Symbolic Logic and Intelligent Machines} should be read today: as a period work of symbolic logic applied to machinery, or as a still-usable account of embodied, behavior-organizing intelligence. More analytically, the question is whether Berkeley's treatment of logic and machinery stays at the level of formal description, or whether it already specifies a machine architecture in which sensing, state, control, and action are treated as coordinated components of intelligence.

The working thesis is that Berkeley's work is best read as a technically viable account of embodied machine intelligence whose implications have not been exhausted. He inherits formal machinery from Boole, Venn, and Shannon, yet uses that inheritance to theorize intelligent systems as physically realized, temporally organized, and behaviorally active machines \citep{boole1847mathematical,boole1854laws,venn1894symbolic,shannon1938symbolic}. Heiserman offers a comparison case in which memory and generalization are layered onto reflexive behavior \citep{heiserman1981robot}. Together, Berkeley and Heiserman identify a descriptive vocabulary that has not been fully tested against current assumptions about robotic intelligence. For this paper, that vocabulary can be stated schematically as a pipeline of sensing, state or memory, control, action, and adaptation. The literature review supports that claim by placing Berkeley among adjacent traditions in formal logic, cybernetics, adaptive machinery, and embodied control.

\section{Literature Review}

The literature relevant to this paper can be organized into four principal bodies plus a methodological note. The first establishes the logical and formal inheritance running from Boole and Venn through Berkeley and into the comparison with Heiserman. The second clarifies the mid-century symbolic-logic context against which Berkeley's distinctiveness becomes visible. The third places Berkeley within the broader cybernetic and machine-intelligence conversation shaped by Wiener, Ashby, Turing, and Walter. The fourth brings that discussion into contact with later embodied and behavior-based robotics. A final methodological note explains how the present paper uses primary-source reconstruction to connect literatures that are usually kept apart. Read in that order, the review moves from logical foundations, to contextual differentiation, to cybernetic placement, to embodied-robotic comparison, and finally to interpretive method.

The first body consists of the primary sources that directly carry the argument: Berkeley, Boole, Venn, Shannon, and Heiserman. Berkeley is the principal object of interpretation, but the point of the paper is not recoverable from Berkeley alone. Boole provides the underlying claim that logical operations can be written in a compact symbolic calculus and manipulated with rule-governed precision \citep{boole1847mathematical,boole1854laws}. Berkeley does not invent his symbolic resources from scratch; he inherits a tradition in which reasoning is already being recast as explicit operation. Boole gives him the confidence that classes, statements, and operations can be made exact enough to serve as components in a design language. Venn matters in two distinct ways. In \emph{Symbolic Logic}, he sharpens the class-logic tradition Berkeley works within and clarifies the status of logical symbolism as a general calculational language rather than a mere classroom mnemonic \citep{venn1894symbolic}. Berkeley's use of symbolic logic as a practical language for machine organization sits more comfortably once that broader Vennian conception is in view. In \emph{The Logic of Chance} and \emph{The Principles of Empirical or Inductive Logic}, meanwhile, Venn moves toward probability, induction, and knowledge under conditions that are not exhausted by strict deduction \citep{venn1866chance,venn1889empirical}. That second Vennian strand helps distinguish Berkeley's mostly rule-bound machine logic from Heiserman's move toward machines that store experience, accumulate success conditions, and generalize from them. Shannon supplies the crucial bridge from logical relations to relay and switching hardware: once logical form is realizable as circuitry, symbolic structure becomes directly architectural rather than merely notational \citep{shannon1938symbolic}. Berkeley's 1952 article ``Algebra in Electronic Design'' shows him actively working on that bridge at the level of concrete design, using Boolean algebra to simplify circuits and to specify the sensing and steering logic of \emph{Squee} \citep{berkeley1952algebra}. His January 1952 \emph{Astounding Science Fiction} article ``Machine ``Intelligence''' is also relevant here, because it shows Berkeley presenting the machine-intelligence problem to a broad popular audience in explicitly operational terms, moving from calculation and stored information toward machines that could in principle handle ideas, recognition, and organized response \citep{berkeley1952machineintelligence}. If Boole and Venn explain Berkeley's formal inheritance, Shannon explains why Berkeley could treat logic as a machine-building discipline rather than only as a notational one, and Berkeley's 1952 articles show him already carrying that possibility both into compact engineering form and into public machine-intelligence discourse. Heiserman, by contrast, is not part of Berkeley's own bibliographic world, but functions as a comparison point that clarifies how the same broad design space can be pushed from explicit control organization toward adaptive and experiential machine intelligence \citep{heiserman1981robot}. His Alpha, Beta, and Gamma levels supply a vocabulary for comparing fixed behavioral organization, learned response, and confidence-bearing generalization without abandoning embodiment. Heiserman therefore does not merely extend Berkeley chronologically; he sharpens the comparison by asking what happens when a state-organized creature must also remember, prefer, and generalize. Taken together, these five primary figures supply the paper's minimal conceptual arc: formal logic, practical symbolism, hardware realization, machine organization, and adaptive extension.

The second body of literature consists of immediate contextual works in symbolic logic, especially Basson and O'Connor's \emph{Introduction to Symbolic Logic} \citep{basson1960introduction}. This source is worth more than a passing mention because it gives a fairly orthodox mid-century picture of what symbolic logic education looked like when centered on the propositional calculus, the algebra of classes, quantificational form, and formal deduction as such. Read against Basson and O'Connor, Berkeley's difference becomes sharper. He is not simply teaching symbolic manipulation, validity, or proof technique. He repeatedly moves symbolic logic outward into circuit design, temporal organization, state transition, memory, and machine behavior. Without a foil like Basson and O'Connor, Berkeley's practical and architectural language can look like a routine extension of symbolic-logic pedagogy. With it, we can see that Berkeley is selecting from the symbolic tradition and reorienting it toward implementation, control, and intelligent machinery. Basson and O'Connor therefore help establish that Berkeley is unusual not because he knows different logical fundamentals, but because he asks those fundamentals to do different work. This contextual contrast is important to the paper's method: it lets the reader distinguish between Berkeley's inherited logical tools and Berkeley's distinctive use of those tools.

The third body of literature is the cybernetic and control literature: Wiener, Ashby, Turing, and Walter \citep{wiener1948cybernetics,ashby1952design,ashby1956introduction,turing1948intelligent,turing1950computing,walter1953living}. It prevents the present argument from sounding idiosyncratic or anachronistic. If Berkeley is read only against formal logic, his concern with control, behavior, and machine organization can appear eccentric or marginal. Read against cybernetics, those same concerns become legible as part of a broader technical conversation about regulation, signaling, adaptation, and machine activity in embodied systems.

Each author contributes something different to that clarification. Wiener contributes the most general postwar frame of communication, control, feedback, and the parallel treatment of animal and machine systems. He helps define the field in which organized signaling and regulation become central engineering questions. In relation to the present paper, Wiener is less a direct source for Berkeley than a way of situating Berkeley's machine concerns within a wider problem space where behavior and control are already central.

Ashby's two books are particularly important for this paper's vocabulary. \emph{Design for a Brain} is directly relevant because it treats adaptation, equilibrium-seeking, and self-organizing behavior as machine problems rather than mysteries outside mechanism \citep{ashby1952design}. \emph{An Introduction to Cybernetics} adds a more explicit language of state, transition, regulation, variety, and machine description \citep{ashby1956introduction}, making it especially useful for comparison with Berkeley's own state-and-event formalism. Ashby is especially important because he and Berkeley share an interest in state-organized systems while differing over emphasis: Ashby tends to abstract toward general regulatory principles, whereas Berkeley keeps pulling the discussion back toward logical specification, circuit organization, and the engineered composition of intelligent machines. The contrast locates Berkeley more precisely. He is not simply another cybernetic theorist of regulation, but one whose preferred idiom remains logical, design-oriented, and close to implementation.

Ashby's importance becomes still clearer once Berkeley's ``Learning from Experience'' passage is taken seriously. At that point Berkeley can no longer be treated as a thinker of explicit control alone. He is reaching, however incompletely, toward machines whose future conduct is altered by prior success, failure, and feedback. That is precisely where Ashby becomes more than contextual background. He supplies the strongest mid-century account of how adaptive order might arise without appealing to magic, anthropomorphic mystery, or contemporary statistical black boxes. In that respect, Ashby functions here as a missing middle pressure point between Berkeley and Heiserman. He does not erase the difference between them, but he makes their continuity more intelligible by showing how regulation, self-correction, and partial self-reorganization could already be conceived as rigorous engineering questions.

Turing enters this literature review for a different reason. In \emph{Intelligent Machinery} and ``Computing Machinery and Intelligence,'' he keeps open the problem of machine intelligence as something that may require organization, education, learning, and developmental method rather than mere fixed calculation \citep{turing1948intelligent,turing1950computing}. His language of discrete machinery, memory capacity, and even unorganized machines gives us a second comparison class in which machine intelligence is treated as buildable, but not primarily through Berkeley's route of symbolic control architecture. Turing is therefore close enough to Berkeley to be illuminating and different enough to be clarifying: Berkeley is more architectonic and control-explicit, Turing more open to training, programming, and developmental organization. That difference shows that early machine-intelligence thought already contained divergent architectural bets. Berkeley's line is not the only one available, but it is one of the clearest for readers interested in explicit structure.

Walter's role is more concrete and embodied. His work on brain mechanisms and autonomous tortoise-like devices shows how relatively simple circuitry can produce environment-facing, apparently purposive behavior \citep{walter1953living}. Walter matters because he demonstrates in physical form what Berkeley often argues in conceptual and architectural terms: that intelligence can appear at the level of organized sensorimotor behavior without requiring disembodied rationality. If Wiener provides a field, Ashby a state vocabulary, and Turing a contrasting developmental orientation, Walter provides a physically memorable demonstration of why embodied control should be taken seriously at all.

Walter also matters because he prevents the paper from drifting into an overly discursive notion of embodiment. Berkeley can sometimes sound as though a sufficiently rich control description would by itself settle the question of machine life. Walter's tortoises answer with hardware on the floor. They wander, orient, seek, avoid, and recover in ways that make the problem of living organization visible in public space. Walter is therefore not merely an anecdotal parallel to Berkeley. He is evidence that the line from circuitry to creature-like conduct was not only imaginable but materially demonstrable. If Ashby gives the argument its adaptive seam, Walter gives it its embodied seam. Together they help explain why the Berkeley material should be read as part of a serious program about living machine intelligence rather than as an eccentric side branch of symbolic logic.

Taken together, these works do not prove Berkeley's argument, but they do show that Berkeley belongs in an active postwar conversation about regulation, autonomy, adaptation, and machine behavior, not only in the narrower history of formal logic. More importantly, they help the reader separate three questions that are often run together: whether intelligence can be mechanized, whether it must be embodied, and whether its organization should be explicit or learned. Berkeley matters in this landscape because he offers a strong answer to the first two questions and a distinctive answer to the third.

The fourth body of literature is behavior-based or embodied-machine comparison, especially Braitenberg and Brooks \citep{braitenberg1984vehicles,brooks1986robust,brooks1991intelligence}. It gives the reader a later and more familiar contrast class. Many contemporary readers approach embodied intelligence through behavior-based robotics, situated action, or anti-representational traditions. Without addressing that lineage, Berkeley's embodied features may either be overlooked or misdescribed.

This comparison body also helps explain why the present paper is an intervention into the way contemporary robotics research is narrated, not just an exercise in historical recovery. Public discussion around current foundation-model robotics can make it seem as though the decisive novelty lies in the very idea that a machine must integrate perception, action, planning, and adaptation across a whole embodied platform \citep{googledeepmind2025geminirobotics,googledeepmind2026geminier}. What is often newer is not the existence of that problem, but the scale, data regime, and implementation substrate at which it is now being pursued. The older literature matters because it keeps us from confusing a new training stack with the first serious encounter with embodied intelligence itself.

These references are used cautiously, but they should not be treated lightly. Braitenberg is valuable because he shows how surprisingly rich behavior can arise from very simple sensorimotor couplings. This keeps the paper from assuming that intelligence must begin at the level of high symbolic representation. Braitenberg's value here is diagnostic: he reminds us that embodied intelligence can emerge from control organization that is behaviorally effective even when it is structurally simple.

Brooks matters because he explicitly argues for situated, layered, behavior-producing architectures that do not depend on heavy centralized representation. The point is not that Berkeley anticipates Brooks in a straightforward way; he does not. Berkeley is much more willing than either Braitenberg or Brooks to preserve an explicit language of logical condition, internal organization, and designed control. Yet the comparison is still productive because it reveals that Berkeley's machines are not static theorem engines. They are systems that sense, transition, remember, and act in time. Brooks thus helps sharpen a negative and a positive point at once: negatively, Berkeley is not simply a forerunner of later behavior-based robotics; positively, he is also not confined to a disembodied symbol-processing picture of intelligence.

The later behavior-based literature therefore does two things at once: it prevents us from mistaking Berkeley for a contemporary behavior-based roboticist, and it prevents us from misclassifying him as a purely disembodied symbolicist. The resulting picture is more exact. Berkeley occupies a middle position in which symbolic organization and embodied control are not adversaries but cooperating layers of machine intelligence. That positioning matters because the paper's claim depends on precisely this middle ground. If the reader misses that point, the Berkeley--Heiserman comparison can collapse in either direction: Berkeley becomes too formal to matter for robotics, or too loosely embodied to matter for logical architecture. This fourth body of literature helps keep both distortions in check.

The methodological point should be stated explicitly. Historical reconstruction and philosophical clarification are among the ways under-registered lines of thought become visible again, and this paper aims to contribute to that methodological task as well as to the substantive Berkeley--Heiserman argument. One reason the connection developed here has been easy to miss is that it falls between disciplinary narratives that are usually kept apart. By reconstructing the line across logic, cybernetics, machine intelligence, and contemporary embodied robotics, the paper is not only making a claim about Berkeley and Heiserman. It is also showing one way to analyze partially hidden continuities in the history of ideas and engineering practice. The present argument therefore does not reject further historical or philosophical work. On the contrary, it invites more of it. What it claims is that primary-source analytic reconstruction can itself be a useful method for recovering lines of thought that standard classifications often leave too vaguely described.

This review also clarifies the paper's novelty claim. In the literature assembled here, Berkeley and Heiserman usually appear in separate conversations: Berkeley in symbolic logic and machine thought, Heiserman in adaptive robot intelligence, and recent state-based robotics in a largely contemporary engineering register. On that basis, the present argument is not offered as a summary of an established Berkeley--Heiserman literature, but as a synthetic claim that emerges by reading these bodies of work together. The point of assembling these four principal literatures is therefore not just to provide background. It is to show why that linkage is interpretable, technically serious, and worth advancing as an architectural claim.

\subsection*{From Interpretation to Engineering}

The argument of this paper concerns design description as well as renewed historical attention for Berkeley and Heiserman. Berkeley describes intelligent machinery in terms of logical conditions embodied in circuitry, organized control, stored information, state transition, and purposive behavior. Heiserman develops a comparison case in which machine creatures retain successful responses and later generalize from them under confidence-sensitive conditions \citep[pp.~18--20]{heiserman1981robot}. Read together, they define a coherent engineering vocabulary in which intelligence is organized through relations among sensing, internal state, control, memory, and outward action. The remainder of the paper treats Berkeley and Heiserman against the common schema introduced above: sensing, state or memory, control, action, and adaptation.

What makes this line interesting now is its systematic coherence. Both writers assume that intelligent machinery must be organized around explicit states, condition-sensitive transitions, sensorimotor coupling, and coordinated internal and outward behavior. In Berkeley, these elements appear in relay and circuit terms, with emphasis on explicit control organization and physically realized behavior programs. In Heiserman, many of the same elements appear through creature experiments that foreground remembered success, response selection, and generalization. Even where the two differ, they differ along an axis of control organization and adaptivity rather than across a simple hardware-versus-software divide. Read together, they define a design space that contemporary AI often treats separately under symbolic reasoning, control, behavior, and learning.

This does not yet prove that their vocabulary offers the best account of modern robotics. It does justify treating it as a technically serious comparison instead of a completed or obsolete episode. The real question is whether explicit state-based forms of behavior organization remain analytically and practically useful in domains where deterministic timing, embodied interaction, modular design, and constrained compute still matter. If they do, then Berkeley and Heiserman become more than historically interesting.

\subsection*{Logic Embodied in Machinery}

One of Berkeley's major contributions is to insist that symbolic logic is a way of designing machinery, not just a way of describing reasoning in abstraction. The bridge here is Shannon. Once Boolean relations can be realized in relay and switching circuits, conditions and inferences no longer remain on paper or in thought alone. They become structures of hardware \citep{shannon1938symbolic}. In the schema used in this paper, this is the control layer stated in logical form and embodied in mechanism.

Berkeley adopts this possibility as fundamental. In his treatment, classes, statements, and logical conditions become design elements for circuits, control systems, and automatic machines. Logic is therefore operational, not simply explanatory. It specifies distinctions a machine can embody, conditions it can test, and sequences it can regulate. Berkeley's interest in symbolic logic is inseparable from his interest in machinery capable of carrying out organized, reason-like operations in the world \citep{berkeley1959symbolic}. At this point in the argument, Berkeley is supplying the formal basis on which sensing inputs, stored states, and action rules can be coordinated within a single machine design. This is one reason the paper insists on reading Berkeley architecturally. Without that shift in emphasis, the transition from logical relation to organized machine behavior can look like an optional application instead of a central feature of the book's project.

\subsection*{Intelligent Machines as Organized Systems}

The strongest evidence for an embodied reading appears in Berkeley's own definitions and examples. He lists a wide range of machines as ``intelligent machines,'' including traffic-light controllers, air-traffic systems, factory control equipment, game-playing machines, reading and recognizing devices, robots, and digital computers \citep[pp.~66--69]{berkeley1959symbolic}. This list defines the target class broadly. Intelligent machinery is not confined to formal problem solving. It includes systems coupled to signals, environments, movement, and regulated action.

Berkeley's January 1952 \emph{Astounding Science Fiction} article helps show that this was not just a textbook classification repeated later in book form \citep{berkeley1952machineintelligence}. There he presents machine intelligence in operational rather than purely formal terms. He begins from capacities such as taking in instructions, storing information, selecting routines or subroutines on the basis of indications, checking tentative answers, and putting out results. He then frames intelligence through Wechsler's language of purposeful action, rational thought, and effective dealing with an environment, explicitly asking what additional sensing organs or ``perceptors'' a machine would need in order to cope with practical situations more richly. Read this way, the article is not only a speculative prelude to the later discussion of learning. It already treats machine intelligence as an organized relation among stored information, branching procedure, sensing capacity, and environment-facing response.

Berkeley then proposes a general architecture shared by almost every intelligent machine. Such a machine, he says, has inputs, outputs, storage or memory, calculating capacity, and a control unit \citep[p.~70]{berkeley1959symbolic}. In the terms of this paper, that architecture already contains the major elements of the proposed schema: sensing through inputs, state or memory through storage, control through the control unit, and action through outputs. Especially important is Berkeley's remark that the control unit may itself be subject to instructions calculated by the machine, allowing the machine in some sense to guide itself \citep[p.~70]{berkeley1959symbolic}. The autonomy is limited, but it means that control is not imposed only from outside; it can be organized internally as part of the machine's own operation. This passage deserves emphasis because it prevents a common misreading. Berkeley is not just listing machine parts. He is specifying a coordination problem: how perception, memory, processing, and directive structure are assembled so that behavior can be produced and maintained.

The traffic-light example in \emph{Symbolic Logic and Intelligent Machines} shows this coordination problem in a particularly useful simple case \citep[pp.~62--66]{berkeley1959symbolic}. Berkeley does not begin with a humanoid robot or a grand theory of mind. He begins with a controller embedded in an intersection, taking in signals from approaching cars, retaining short-term information through relay memory, distinguishing temporally different arrivals, and issuing an outward response that regulates movement in the world. The example matters because it demonstrates embodiment at a modest but unmistakable level. Berkeley's intelligent machine is already environment-coupled, stateful across time, and organized around appropriate action rather than around disembodied inference alone. That makes the later move to robots cumulative rather than abrupt.

That architectural description becomes more concrete in \emph{The Construction of Living Robots}. There Berkeley does not stay at the level of a general parts list. He proposes a specific ``robot world'' populated by four small robots, food areas, a storehouse, a repair shop, and a birth factory, and he assigns each component a determinate behavioral role \citep{berkeley1952livingrobots}. The storehouse receives food, the repair shop restores damaged robots, the birth factory issues replacements when a robot's identifying light disappears, and each robot is organized to hunt for food when well, deliver it to the storehouse, seek repair when sick, and back away from obstacles. The important point is the design structure: self-preservation, self-maintenance, and reproduction are decomposed into an environment, a set of functional stations, and a recurrent program of condition-governed acts. Berkeley is not just asserting that living-like behavior could exist in hardware. He is sketching an explicit ecological control architecture in which the behavior of each machine depends on organized relations among sensing, internal condition, support infrastructure, and action.

That point deserves to be stated as strongly as the source allows. Berkeley's claim about ``living'' robots is not a casual metaphor attached to ordinary automation. He first asks what would count as life even in the difficult case of an unfamiliar being, then proposes six operative properties: self, sensation and response, death, self-preservation, self-maintenance, and reproduction \citep{berkeley1952livingrobots}. The key move is that the last three properties are not treated as mystical thresholds but as engineering problems. A living machine is one whose ordinary responses tend to avoid destruction, whose organization permits repair and replenishment by drawing on its environment, and whose world includes means by which more machines of its own type can be produced. On this view, ``living'' machine intelligence is not exhausted by inference, planning, or even local adaptation. It concerns the full closure of a behaving system within a world: the machine must be able to persist, recover, and continue its kind through an organized relation among body, environment, support structure, and control.

\subsection*{States, Events, and Temporally Extended Behavior}

If Berkeley had stopped with Boolean algebra and machine architecture, the case for an embodied reading would remain incomplete. It becomes stronger in his treatment of the algebra of states and events. There Berkeley explicitly argues that Boolean calculus must be extended in order to come ``into closer relation with the real world,'' since classes and statements change with time and since machine operation involves changing states and events rather than static truths alone \citep[pp.~145--147]{berkeley1959symbolic}. This is the point at which the schema acquires an explicitly temporal account of state transition.

A state lasts for a time; an event is brief. This distinction allows Berkeley to formalize persistence, interruption, transition, delay, and sequence. He also insists on the practical importance of time units, step-functions, and even intervals in which a machine state is temporarily undefined during transition \citep[pp.~145--147]{berkeley1959symbolic}. Once machine activity is represented as passage through states and events, intelligent behavior becomes describable as organized temporal process rather than as a mere arrangement of timeless logical forms. In the present argument, this is the component that connects stored condition, control logic, and temporally ordered action.

The transcription of \emph{The Construction of Living Robots} shows that Berkeley pressed this temporal vocabulary into a much more operational form than a casual reading of the 1959 book might suggest. In the robot-species design section, he enumerates eleven needed senses for the robot, including left-eye and right-eye light detection, food detection, storehouse and repair-shop detection, obstacle detection, and the compound conditions of ``becoming sick'' and ``becoming well'' \citep{berkeley1952livingrobots}. He then writes the program in relay terms using delayed variables, hold contacts, state relays, and timed transitions, culminating in a topology of states such as seeking food, seeking the storehouse, putting food down, seeking the repair shop, and backing up from obstacles. This matters for the present paper because it confirms that Berkeley's language of states and events was not philosophical gloss. He was treating temporally extended behavior as something that could be explicitly programmed in circuit form through persistence conditions, event triggers, and delay operators.

This temporal organization is also one place where Berkeley comes unexpectedly close to later embodied-cognitive concerns. His robot is not just a reactive machine that maps a stimulus to a motor output. It dwells in states, maintains them across time, exits them when events intervene, and re-enters them when the world changes again. Hunger analogues, obstacle encounters, sickness, repair, food transport, and return-to-motion are all treated as parts of one continuing organization. Modern embodied robotics often discusses such matters under the language of situated control, hybrid systems, sensorimotor loops, maintenance, recovery, and task persistence. Berkeley's vocabulary is different, but the underlying issue is recognizably the same: how a physically realized agent preserves coherent conduct across interruption, damage, replenishment, and changing environmental conditions.

\subsection*{Robots and Behavioral Programs}

Berkeley's robot chapters make the behavioral implications even clearer. He defines a robot as a machine able to ``behave'' by itself, taking in sensations by means of sensing devices, performing actions by means of acting devices, and correlating sensations and actions by means of circuits or mechanisms expressing instructions \citep[pp.~177--178]{berkeley1959symbolic}. This definition explicitly links sensing, control, and action. The robot is not presented as a proof engine with motors attached. It is presented as a behaving system whose intelligence lies in the organization of sensation, action, and instruction. That formulation matters because it shifts the object of analysis from isolated inference to organized conduct. Berkeley's broader robot writing reinforces the same point. In \emph{The Construction of Living Robots}, he states that the purpose is to discuss the properties of living beings and outline how to construct robots from hardware that exhibit essential living behavior \citep{berkeley1952livingrobots}. Once that shift is made, Berkeley's robot examples cease to look like colorful supplements and start to look like direct evidence for the paper's architectural reading.

His examples are equally revealing. Berkeley's electronic robot squirrel ``Squee'' is organized around activities such as hunting, homing, and depositing, each of which persists until relevant conditions terminate it \citep[pp.~177--180]{berkeley1959symbolic}. In the earlier \emph{Radio-Electronics} article ``Light Sensitive ... Electronic Beast,'' Berkeley had already described `Squee` in terms of sensing organs, acting organs, and an electronic relay brain, explicitly presenting Boolean algebra as the design method coordinating those parts \citep[pp.~46--47]{berkeley1951electronicbeast}. The surviving construction plans push that point from conceptual description toward hardware layout: they break `Squee` down into separate right-eye and left-eye amplifier circuits, dedicated scoop and steering motor controls, a steering-column assembly, explicit electrical and mechanical parts lists, and a short operating procedure for bringing the whole machine online \citep{berkeley1952squeeconstruction}. Taken together, these materials show that Berkeley was not invoking creature-like behavior only at the level of metaphor or popular exposition. He was also specifying a concrete sensorimotor assembly whose behavioral repertoire depended on coordinated sensing, relay logic, actuation, and startup procedure. Likewise, the ``robot animal'' problem is introduced explicitly as a programming problem in which one must assign a number of states to a robot and then specify changes from one state to another when certain events happen \citep[pp.~177, 180--181]{berkeley1959symbolic}. The 1952 living-robots report makes the same point with still greater specificity: the robot's actions are narrowed to forward or backward driving, steering, and scoop movement, while the program is organized around explicit state transitions such as finding food, arriving at the storehouse, becoming sick, seeking repair, and resuming forward motion after a timed backing-up interval \citep{berkeley1952livingrobots}. These are behavior-organizing schemes in which logical conditions select, maintain, and redirect activity. In the shared schema, they are explicit instances of sensing coupled to state, control, and action over time.

\subsection*{Berkeley and Heiserman}

The comparison with David Heiserman clarifies both Berkeley's strengths and his limits. Heiserman's \emph{Robot Intelligence ... with Experiments} presents an evolutionary hierarchy in which Alpha-class creatures make random reflexive responses to conditions in the immediate environment, Beta-class creatures add a memory of past successful responses, and Gamma-class creatures generalize from prior experience so that they can deal more effectively with situations not yet encountered \citep[pp.~17--20]{heiserman1981robot}. In Heiserman's own construction, Beta is not a break from Alpha but an Alpha-class mechanism equipped with memory, while Gamma adds the ability to project beyond the immediately given case through generalization \citep[pp.~19--20]{heiserman1981robot}. In that framework, machine intelligence is lived by a creature in an environment: it senses, acts, remembers, and, in the Gamma case, anticipates. This matters for the present paper because it lets us compare Berkeley and Heiserman at the level of architecture instead of reputation. Both are concerned with machines whose behavior depends on structured relations among input, internal organization, stored information, and action, but Heiserman specifies more clearly than Berkeley how reflex, memory, and generalized expectation are layered into one developmental hierarchy.

Berkeley does not reach this level of adaptivity. His machines are mostly engineered top-down rather than self-modifying through reinforcement, stored success conditions, or confidence-based generalization. He does not provide Beta-style memory schemes in which workable responses are filed under recurring conditions, nor Gamma-style extension from solved cases to similar unsolved ones \citep[pp.~19--20]{heiserman1981robot}. Even so, the continuity remains important. Berkeley already supplies the control-and-behavior side of the story: intelligence as physically realized organization, behavior as transitions among activities, sensing and acting as core capacities, and internal control as the basis of outward competence. Heiserman does not replace that framework so much as deepen it by showing one concrete developmental sequence through which reflexive behavior can become memory-guided and then generalized. This is the hinge of the comparison. Berkeley supplies an architecture of embodied control; Heiserman turns that architecture toward memory-guided adaptation and anticipatory extension.

Yet that contrast must now be made more carefully than before, because Berkeley's January 1952 \emph{Astounding Science Fiction} article shows that he had already begun to think directly about learning from experience as a machine problem \citep{berkeley1952machineintelligence}. In the section explicitly titled ``Learning from Experience,'' Berkeley does not stop at the general claim that an intelligent machine might someday improve. He asks how an animal problem-box test from comparative psychology could be translated into machine form. His answer is strikingly operational. The machine is to be given registers corresponding to places in the problem box, digits encoding whether food is present, whether it remains locked, and whether a place contains one of the special plates, together with comparing circuits and subroutines for finding food, moving among places, selecting where to go next, and preferring one subroutine over another when it yields fewer unsuccessful trips. In other words, Berkeley is no longer speaking only of logic embodied in circuitry or of a robot ecology organized for persistence. He is describing, in a distinctly early and still partly analog language, a proposed machine whose later conduct is meant to be altered by what happened in earlier trials. The idiom is not Heiserman's, and the architecture remains more sketched than demonstrated, but the direction is unmistakably toward experience-structured control.

This matters because it sharpens the exact point of continuity and difference between the two authors. Berkeley still does not supply anything as explicit as Heiserman's Alpha, Beta, and Gamma hierarchy. He does not yet differentiate with Heiserman's precision between present-tense reflex, remembered successful response, and generalization from successful cases to novel but similar ones. Nor does he give a creature-level account of confidence-bearing revision. What he does show is that the adaptive problem was already visible to him. The last pages of ``Machine ``Intelligence''' make clear that Berkeley understood learning as a design issue involving coded world conditions, retained machine-accessible state, selection among alternative procedures, and the possibility that very general negative-feedback instructions might allow a machine to ``work out'' part of its own effective organization. His appeal to Ashby is important at this limited level. Berkeley is not yet presenting a demonstrated adaptive architecture; he is identifying self-modifying control as a genuine engineering prospect and sketching one way such a prospect might be approached within a structured environment.

More precisely, Berkeley's 1952 ``life species'' scheme already covers much of the total problem-space that Heiserman later occupies. Berkeley specifies a bounded world, multiple agents, differentiated environmental roles, food or energy analogues, repair, replacement, obstacle avoidance, explicit sensing channels, explicit action channels, and a temporally organized program that moves the creature among named conditions such as seeking food, finding the storehouse, becoming sick, seeking repair, and resuming motion \citep{berkeley1952livingrobots}. In that sense Berkeley offers a broad ecology of machine life instead of a single isolated task robot. What Heiserman adds is a much more detailed account of how response quality changes with experience. Alpha creatures respond reflexively to present conditions, Beta creatures couple that behavior to stored successful responses, and Gamma creatures generalize from successful cases to similar but previously unencountered situations \citep[pp.~17--20]{heiserman1981robot}. Berkeley therefore encompasses a surprisingly large outer shell of the creature problem, while Heiserman works inward on the memory and inference mechanisms by which a creature changes its future conduct.

This sharper comparison also helps prevent two opposite misreadings. One is to treat Berkeley as if he offered only static circuitry while Heiserman alone supplied creature intelligence. The other is to claim that Berkeley had already solved the adaptive problem in advance. Neither is satisfactory. Berkeley already has world structure, sensorimotor coupling, state transition, maintenance, and reproduction in view; Heiserman does not need to invent those themes from nothing. But Heiserman's central contribution lies elsewhere: he differentiates classes of creature by their relation to experience itself, moving from present-tense response, to remembered success, to generalized expectation, and eventually to confidence-weighted revision. The relation between the two writers is therefore neither simple anticipation nor simple discontinuity. It is better understood as overlap in the ecological and control frame, together with a real asymmetry in the treatment of learning, memory, and anticipatory inference.

The correspondence becomes even more instructive when one looks at what each author chooses to internalize and what each leaves in the environment. Berkeley externalizes a great deal of the life-support problem. His robots depend upon a storehouse, a repair shop, and a birth factory; the world is deliberately organized so that self-preservation, self-maintenance, and reproduction can be achieved through recurrent access to those stations. Heiserman, by contrast, keeps the world comparatively austere and works instead on the creature's internal relation to it. His Alpha creatures avoid barriers or pursue beneficial conditions in the moment; his Beta creatures remember which responses have worked; his Gamma creatures generalize successful patterns to unencountered cases; and later Delta formulations assign confidence levels and seek circumstances in which low-confidence responses can be tested. Berkeley therefore spreads ``life'' across an ecology of agent plus support environment, while Heiserman concentrates on the internal modification of response within the creature itself. Both are theories of living machine intelligence, but they articulate different loci of organization: Berkeley at the level of ecological closure and explicit behavioral programming, Heiserman at the level of experiential revision, habit, conjecture, and confidence.

That difference matters for embodied cognitive robotics today because it surfaces two questions that are often collapsed into one. The first is whether an intelligent machine has enough sensorimotor and control organization to persist meaningfully in a world at all. The second is whether that machine can improve, extend, or revise its conduct on the basis of prior experience. Berkeley attacks the first question much more directly than many later writers do: he asks what a machine would need in order to count as a living, persisting, replenishing, reproducing being in an ordinary environment. Heiserman attacks the second much more directly than Berkeley does: he asks how a creature already situated in a world can move from random response to memory-guided response and then to generalized expectation. Read side by side, the two writers do not merely share a concern with embodied control. They divide between them a surprisingly large portion of what later embodied cognitive robotics still has to explain: persistence, maintenance, world-coupling, action selection, memory, anticipation, and revision.

Stated in the strongest comparative form, ``living'' machine intelligence in this paper does not mean disembodied symbol manipulation supplemented by sensors, nor does it mean mere locomotion or reactivity. It means an organized capacity to remain behaviorally coherent within a changing world. Such coherence has several interlocking dimensions. A creature must discriminate conditions through sensation; it must preserve enough internal continuity to remain the same agent across time; it must select and sustain actions rather than merely emit isolated responses; it must cope with interruption, damage, depletion, and renewed opportunity; and at the richer end it must let prior success alter future conduct. Modern embodied cognitive robotics often distributes these requirements across different technical vocabularies such as homeostasis, viability, situated control, behavioral arbitration, active perception, memory, learning, and predictive adjustment. The value of the Berkeley--Heiserman comparison is that it lets many of those concerns be seen in a simpler and more historically transparent layout. Berkeley emphasizes viability, maintenance, and organized world-dependence. Heiserman emphasizes adaptive retention, generalization, and the progressive enrichment of conduct. Together they show that a machine begins to look ``alive'' not when it passes a verbal test, but when sensing, state, action, support, and experience are integrated tightly enough that the agent can continue, recover, and alter its way of going on.

The difference, then, is not that Berkeley lacks embodiment; his machines are explicitly sensorimotor, state-organized, and physically realized. The difference is that Berkeley usually presents intelligence as engineered control organization, whereas Heiserman presents a creature architecture that is further differentiated by remembered success and generalization. Their relation is therefore better understood as architectural continuity with a sharper developmental elaboration in Heiserman.

\section{A Reconstructed Design Vocabulary}

The paper's main constructive contribution is to restate the Berkeley--Heiserman comparison as a reconstructed design vocabulary instead of a historical association. The claim is that the organization recovered from the earlier texts is specific enough to support technical description and comparison: behavior can be organized through explicit relations among sensing, state retention, control, action, recovery, and, in Heiserman's stronger form, adaptation. A companion 2026 study by the present author, \emph{A Behavioral State Vocabulary in Sony ERS-111 R-CODE}, studies Sony's preserved \texttt{R-CODE} sample corpus for the \texttt{ERS-111} AIBO and argues that many apparently distinct routines are organized by a compact recurring behavior vocabulary \citep{tucker2026behavioral}. The function of that companion paper here is not to serve as independent confirmation or to displace the paper's main historical and conceptual contribution with a Sony-centered one. It develops, in a constrained robotic corpus, a concrete analytic application of the same line of thought. Its importance for the present argument is demonstrative rather than decisive: it provides a later setting in which behavior organization can be analyzed, compared, and described at the level of design.

That reconstructed vocabulary should be understood in a slightly stronger sense than mere state organization. If the Berkeley material ended only with explicit control, states, and behavior programs, then the AIBO comparison would mainly concern explicit embodied organization. But Berkeley's ``Learning from Experience'' discussion suggests a further threshold: explicit state-organized control can at least be rendered available for revision once conditions, outcomes, and procedure selection are made machine-explicit. Section 4 therefore does not inherit one tradition of fixed Berkeleyan control and then tack on a separate Heiserman layer from outside. It works at the junction between them, where organized embodied behavior becomes a plausible substrate for modification by prior success and failure even before full creature-style generalization has been achieved. That junction is one of the paper's most important claims, because it identifies a level at which embodied control and adaptive revision need not be treated as alien paradigms, while still keeping clear that Berkeley mostly marks this possibility in sketch form rather than demonstrating it in a mature adaptive design.

In the terms developed earlier in this paper, the contribution of the \texttt{R-CODE} analysis is to make the common schema empirically explicit within one preserved robotic corpus. Across the sample set, routines can be read as structured combinations of initialization, sensing, state retention, decision, action, synchronization, recovery, and repetition rather than as flat command sequences. That is already close to Berkeley's machine description. Inputs, outputs, storage, control, and behavior appear here in practical script form. It also sharpens the comparison with Heiserman, because once such structures are made explicit, one can ask which parts are fixed, which are mode-based, which are responsive to sensed conditions, and which might plausibly support adaptive revision.

Just as importantly for later engineering work, this schema does not require a monolithic implementation. Once behavior is decomposed into explicit sensing, local state, transition conditions, action routines, and supervisory mode structure, the same organization can in principle be distributed across cooperating components that run at different cadences. Fast reactive loops may remain close to sensors and actuators, while slower supervisory or revisory processes coordinate larger behavioral shifts. In that respect, the Berkeley--Heiserman comparison is compatible with later embodied implementations without implying any single necessary destination for them.

The strongest value of this use-case lies in the shift from single-script reading to corpus-level abstraction. The claim of the companion study is that many sample programs share a small embodied grammar centered on startup conditions, environmental or bodily sensing, conditional branching, iterative motor activity, synchronization, and recovery. The updated corpus aggregate makes that claim more concrete: across 54 diagrams, Tucker counts 292 total state instances organized by 47 unique state titles. The point of those totals is descriptive, but not only descriptive. They show that visible behavioral variety is repeatedly produced from a relatively compact control vocabulary. In practical terms, this means that behaviors such as locomotion demos, obstacle-avoidance routines, and ball-tracking scripts can be compared within one analytical frame instead of treated as isolated scripts. That move matters for the present paper because it supplies exactly the kind of intermediate level of description that Berkeley's state-and-event formalism calls for: not raw machine code, not disembodied symbolism, but organized behavioral structure.

The same corpus view also sharpens the architectural claim. Tucker's most frequent state titles include \texttt{Sense / Decide}, \texttt{Action Loop}, \texttt{Boot / Safe Pose}, \texttt{Synchronize}, \texttt{Sense Fall State}, and \texttt{Recover}. Read together, they make explicit a recurring embodied workflow that can be summarized as initialize, sense, decide, act, synchronize, and repeat. That sequence is especially useful for the present paper because it renders Berkeley's own language of inputs, control, states, events, and outputs in a compact behavioral form. It also clarifies why the AIBO corpus is relevant to Heiserman: once such recurring loops are explicit as reusable units, one can ask how they might be layered, revised, or compared with more adaptive creature-like organization.

The \texttt{R-CODE} corpus is especially useful because it spans a graded range of behavioral complexity and now makes that grading explicit. At the simpler end are activation or playback-like routines that largely sequence prepared actions. More complex cases add persistent monitoring and bodily recovery, as in locomotion scripts that watch fall state and trigger restorative behavior. Tucker's updated ladder then inserts a distinct intermediate rung of contact-conditioned response, where the shared control backbone is redirected toward expressive or social interaction rather than locomotion alone. Still richer cases add local branching, environmental comparison, and eventual mode-governed task control, as in maze navigation and football behaviors. This ladder of examples makes it possible to observe, within one preserved platform, a progression from startup regularization, to monitored action, to interaction branching, to decision loops, to fuller state-governed behavioral organization. That progression is directly relevant to the present argument, because it shows how explicit states, sensory conditions, control branches, and recurrent actions combine into a usable descriptive vocabulary for embodied intelligence.

Two brief examples make the point more tangible. A script such as \texttt{Move} can be read at the behavioral level as: boot, assume a safe pose, repeat forward motion, monitor fall state, and invoke recovery if needed. In the extracted analysis, this routine contains four states and five transitions, and its sensed variable is the fall indicator \texttt{Gsensor\_status}. In the paper's schema, that is already a nontrivial organization of action, sensing, control, and recovery instead of a bare motion sequence. The script does not simply command the robot to walk. It couples repeated action to bodily monitoring and to a distinct restorative path, which means that the behavior is organized around conditional continuity instead of one uninterrupted motor command.

A richer case such as \texttt{Football} makes the state structure more explicit. At the behavioral level, the routine can be summarized as search, detect ball, pursue, kick, and resume, with fall recovery and tracking loss handled as separate branches. The local extract identifies thirty-one states, fifty-four transitions, sensed variables including \texttt{Cdt\_npixel}, \texttt{Gsensor\_status}, \texttt{Head\_pan}, \texttt{Head\_tilt}, and \texttt{Psd\_range}, and explicit internal variables such as \texttt{mode}, \texttt{head}, and \texttt{lost}. The important point here is the mode structure underneath the surface task. Search and pursuit are not just different motions. They are distinct behavioral states linked by sensor conditions, maintained by internal variables, and re-entered when the environment changes. This is precisely the level at which the paper's Berkeley--Heiserman claim becomes concrete: the routine is not well described as a bag of commands, but as an organized cycle in which sensing, temporary state, branching, action, and recovery cooperate over time.

The corpus becomes even more informative when several scripts are compared together. \texttt{C-}\allowbreak\texttt{Tracking} is close to capability activation: a minimal startup-and-act pattern. \texttt{Move} adds explicit monitoring and recovery. \texttt{Contact Response} shows how the same shared backbone can be redirected toward local interaction branching. \texttt{Maze} marks the important intermediate threshold at which the behavior becomes visibly decision-structured, because it repeatedly samples local conditions, compares alternatives, and reorients under environmental constraint. \texttt{Football} extends that same logic into persistent mode structure and a fuller embodied state machine. Presented in that order, the sample set yields more than isolated examples. It yields a compact comparison ladder in which one can observe increasing behavioral organization from simple activation, to monitored action, to interaction branching, to decision loops, to explicit mode-based control. That progression is itself part of the paper's contribution, because it makes the architecture visible as a graded design vocabulary rather than as a single all-or-nothing theory.

\begin{center}
\small
\setlength{\tabcolsep}{4pt}
\begin{tabularx}{\textwidth}{@{}lcc>{\raggedright\arraybackslash}X>{\raggedright\arraybackslash}X>{\raggedright\arraybackslash}X@{}}
\hline
Script & States & Transitions & Key Sensing & Recovery & Control Form \\
\hline
\texttt{C-Tracking} & 1 & 0--1 implicit & none explicit & none & triggered capability activation \\
\texttt{Move} & 4 & 5 & \texttt{Gsensor\_}\allowbreak\texttt{status} & explicit & monitored recovery loop \\
\texttt{Contact Response} & 4--5 & local branches & contact cue & none explicit & contact-conditioned branch loop \\
\texttt{Maze} & 12 & multi-branch & \texttt{Distance}, \texttt{Head\_pan} & retry/escape & environmental decision loop \\
\texttt{Football} & 31 & 54 & vision, posture, range & explicit & mode-governed pursuit cycle \\
\hline
\end{tabularx}
\end{center}

Even in this compressed form, the comparison makes a concrete point that is easy to miss in script-by-script reading: the corpus is not just varied in theme, but ordered in control and flow complexity. The progression from one state and little or no explicit transition structure in \texttt{C-Tracking}, to the five-transition monitored loop of \texttt{Move}, to the contact-conditioned branch logic of \texttt{Contact Response}, to the search-and-reorientation logic of \texttt{Maze}, to the fifty-four-transition, mode-governed pursuit cycle of \texttt{Football}, gives the reader a visible scale along which the behavior vocabulary grows.

That pedagogical point should be stated explicitly. For readers encountering this material for the first time, the jump from a compact monitored routine such as \texttt{Move} to a dense mode-driven routine such as \texttt{Football} can be too abrupt if the intermediate cases are not given enough space. The present paper therefore uses the ladder not only as evidence but also as a reading sequence. In that sequence, \texttt{Contact Response} and \texttt{Maze} do the crucial bridging work: the former shows how the common loop branches around interaction, while the latter shows how monitored action becomes decision-structured behavior before the reader is asked to absorb the richer mode logic of \texttt{Football}. A fuller companion treatment of the AIBO corpus can extend this progression further, but the present sequence already shows how the method of analysis is learned through increasingly rich behavior blocks rather than by a leap between simple and highly structured examples.

The point here is not to claim direct historical descent from Berkeley to Sony or from Heiserman to \texttt{R-CODE}. Nor is it to treat one paper by the present author as independent confirmation of another. The evidential function of the section is narrower and still useful: it shows that the ingredients isolated in the earlier parts of the paper remain available as engineering abstractions in a contemporary analysis context. When the AIBO corpus is read through a behavioral state vocabulary, Berkeley's emphasis on temporally organized machine activity becomes explicit again, and Heiserman's emphasis on creature-level organization becomes easier to operationalize as a comparison problem instead of a separate conceptual universe.

If a preserved constrained robotic corpus can be redescribed in terms of stable behavior blocks, recurring state transitions, recovery paths, and mode-based embodied control, then the Berkeley--Heiserman comparison is not only of historical interest. It becomes a tractable basis for comparative reconstruction and technical experimentation. One can analyze existing behaviors, compare them across a corpus, extract a reusable control vocabulary, and then ask which parts of that vocabulary remain useful for describing or building embodied systems. Tucker's updated discussion pushes that point beyond analysis toward construction: once the shared loop is explicit, it can be treated as a library of reusable design units for new native robotic systems instead of only a retrospective description of AIBO code. On that basis, the present reconstruction does more than illustrate relevance. It supplies a methodological bridge between historical interpretation and technical reconstruction. It also stands in a substantive relation to later embodied-control literatures that likewise treat organized sensorimotor structure as a serious engineering object \citep{brooks1986robust,brooks1991intelligence,braitenberg1984vehicles}. More specifically, it supports later work in which fast reactive behavior, slower supervisory control, and distributed implementation are treated as alternative ways of realizing one explicit behavioral organization instead of disconnected design choices.

\section{Conclusion}
\enlargethispage{\baselineskip}

Berkeley's contribution can be stated in technical terms. He treats intelligent machinery as an
organized relation among inputs, outputs, storage, calculation, control, states, events, and
behavioral programs. On that basis, Berkeley is more than a symbolic logician applying formal
methods to machines. He provides a control-oriented account of how intelligent conduct can be
specified, embodied, and temporally organized in machinery.

The comparison with Heiserman sharpens both the reach and the limit of that account. Berkeley
does not formulate adaptive, reinforcement-based, or confidence-driven machine intelligence in the
Heiserman manner. What he does provide is the groundwork: sensing, internal state, control, and
action organized in explicit machine form. Heiserman extends that framework by developing the
adaptive component more fully through remembered success, generalization, and creature-level
learning behavior.

Berkeley and Heiserman together illuminate a substantial part of the total problem of machine
intelligence. They define a vocabulary in which logic, explicit state organization, embodied
control, recurrent behavioral organization, and adaptive response can be treated as compatible
components of machine intelligence. Contemporary work on behavioral state vocabularies for
constrained robotic systems supports the claim that this framework remains analytically useful
\citep{tucker2026behavioral}. In the present paper, that later robotic material serves a supporting
demonstrative role within the larger contribution: the reconstruction of an account of living
embodied intelligence and its use in evaluating current robotics discourse. The framework now warrants
systematic evaluation along at least
three concrete dimensions: whether a common behavioral vocabulary transfers across multiple robotic
corpora, whether such vocabularies improve modular design and behavior comparison in practice, and
whether explicitly state-organized systems can be compared fruitfully with more adaptive forms of
revision without losing explanatory and engineering clarity. Those questions would test the scope
of the thesis across additional robotic corpora and comparative settings.

This also clarifies the paper's intervention into present robotics discourse. If contemporary language-model and vision-language-action systems are taken as the leading edge of embodied AI, then the question is whether successful behavior alone is enough to count as an adequate account of controlled embodied intelligence. The argument developed here says that it is not. A robot may follow instructions, integrate multiple sensory channels, and even improve by iterative trial, while still leaving underdescribed the organization that makes its conduct stable, recoverable, maintainable, and intelligible as a living machine process. Berkeley contributes a language of explicit control architecture, temporally organized behavior, and ecological persistence. Walter shows that circuitry can yield creature-like world engagement. Ashby contributes a theory of adaptive regulation and self-correction. Heiserman contributes an account in which memory, generalization, and confidence alter future conduct. Taken together, they form a vocabulary against which present systems can be assessed. On that basis, the paper does not claim that contemporary LLM-driven robotics is misguided or unimportant. It claims something more precise: that the dominant current paradigm still often demonstrates capability without an equally explicit theory of controlled, embodied, and developmentally structured machine intelligence.

One engineering implication should be stated plainly. If the
Berkeley--Heiserman comparison isolates a useful behavioral vocabulary, then embodied intelligence
need not be implemented as a single undifferentiated control mass. It can instead be organized as
cooperating layers or cadences: fast local reactive processes, slower supervisory reorganization,
and adaptive revision operating over a shared behavioral vocabulary. Such an arrangement need not
collapse into one executable center; it can be instantiated across coordinated components so long
as the behavioral organization remains explicit across them. The paper provides conceptual groundwork for later attempts to realize such
organization on contemporary distributed substrates.

\bibliographystyle{unsrtnat}
\bibliography{references}

\end{document}